\documentclass[11pt,a4paper]{article}
\usepackage[hyperref]{eacl2021}
\usepackage{times}
\usepackage{latexsym}

\usepackage[linesnumbered,ruled,vlined]{algorithm2e}
\usepackage{multirow}

\usepackage{microtype}

\aclfinalcopy 

\title{Short Text Clustering with Transformers}

\author{Leonid Pugachev \\
  Moscow Institute of \\
  Physics and Technology \\\
  \texttt{leonid.pugachev@phystech.edu} \\\And
  Mikhail Burtsev \\
  Moscow Institute of \\
  Physics and Technology \\}

\date{}

\begin{document}
\maketitle
\begin{abstract}
Recent techniques for the task of short text clustering often rely on word embeddings as a transfer learning component. This paper shows that sentence vector representations from  Transformers in conjunction with different clustering methods can be successfully applied to address the task. Furthermore, we demonstrate that the algorithm of enhancement of clustering via iterative classification can further improve initial clustering performance with different classifiers, including those based on pre-trained Transformer language models.

\end{abstract}

\section{Introduction}
There are currently a lot of techniques developed for short text clustering (STC), including topic models and neural networks. The most recent and successful approaches leverage transfer learning through the use of pre-trained word embeddings. In this work, we show that high quality for STC on the range of datasets can be achieved with modern sentence level transfer learning techniques as well. We use deep sentence representations obtained using the Universal Sentence Encoder (USE) \cite{DBLP:journals/corr/abs-1803-11175, DBLP:journals/corr/abs-1907-04307}.

Training of deep architectures can be effective for particular clustering tasks as well. However, application of deep models to clustering directly is difficult since we do not have labels a priori. We show that fine-tuning of classifiers such as BERT \cite{devlin2018bert} and RoBERTa \cite{liu2019roberta} for clustering can be done with the Enhancement of Clustering by Iterative Classification (ECIC) algorithm \cite{rakib2020enhancement}. Thus, we develop a combined approach to STC, which benefits from the usage of deep sentence representations obtained using USE and fine-tuning of Transformer models.

The main contributions of the work are as follows. First, we demonstrate that sentence level transfer learning for clustering which has not been a common technique so far gives good results. Second, fine-tuning of deep models for clustering is hindered because of the lack of labeled data and we propose to use the ECIC algorithm with deep models which has not been done before to tackle this problem. Third, we analyzed different combinations of components as constitutional parts of the algorithm, tested different schemes to handle weights during fine-tuning over iterations and developed a new stopping criterion for the algorithm.

\section{Related work}
One major direction in STC is based on Dirichlet multinomial mixture topic models \cite{yin2014dirichlet,jipeng2019short} including GSDPMM \cite{yin2016model}. Some variants of these models incorporate word embeddings \cite{nguyen2015improving,li2017enhancing,jipeng2019short}. These models assume that each document contains only one or a few topics. The models have several advantages over conventional topic modeling such as latent Dirichlet allocation, when used for short texts. First, they better cope with the sparseness of short texts, which carry limited information about word co-occurrences. Second, these models can automatically infer the number of topics. Since only one topic is presented for each document, it is straightforward to use these topic models for clustering, assuming all documents with the same topic as belonging to the same cluster.

Recent works have considered a neural approach for STC. In \cite{xu2015short, xu2017self}, authors propose to encode texts by pre-trained binary codes. Embeddings of words are then fed in the convolutional neural network which is trained to fit the binary codes. Finally, the obtained representations are used as features with $k$-means clustering algorithm. The work of \cite{hadifar2019self} uses a somewhat similar strategy called Self-Taught Approach (STA). An autoencoder is pre-trained to obtain low-dimensional features and then learn it together with clustering algorithm by iteratively updating the weights of the autoencoder and centroids of clusters. Finally, they use the resulting features with $k$-means clustering algorithm. Another idea is to use attentive representation learning with adversarial training for STC \cite{zhang2019attentive}. The work of \cite{rakib2020enhancement} sets the state-of-the-art results on the range of short text datasets using the ECIC algorithm which is simpler than in \cite{hadifar2019self}. They use averaged word embeddings as features for short texts and clustering algorithms such as $k$-means, to get the initial label assignment. The clustering performance is then improved with iterative outlier detection and classification.

\section{Model}
In our work, we made several important modifications to the ECIC algorithm\cite{rakib2020enhancement} to improve their results. Namely, we included modern deep learning components such as USE, BERT and RoBERTa in the algorithm as well tested various methods to handle weights during fine-tuning over iterations such as resumption and re-initialization and developed a new stopping criterion for the algorithm.  The general outline of the algorithm is shown in Algorithm \ref{enh}. At the initial stage, clustering is carried out using one of the widely used clustering methods (see below). An algorithm for outlier detection is then used to split the dataset into train and test parts. Additional samples can be moved from the train to the test set based on the $P$ number sampled randomly in the range from $P_1$ to $P_2$. The train part is used to train the classifier. Outliers and some number of the additional samples are used as a test set and predictions for the test set are used to relabel the dataset. Steps with outlier detection, classification, and relabeling are then repeated until the stopping criterion is reached or the maximum number of iterations is exceeded. As will be shown below, this iterative procedure leads to improved clustering results in many cases.

\begin{algorithm}
\label{enh}
\SetAlgoLined
\KwResult{Enhancement of Clustering}
 Dataset $D$ with $N$ texts and $K$ clusters\;
 Apply initial clustering and labeling $L$\;
 Set the number of iterations $T$\;
 \While{$j\le T$ and the stopping criterion $\delta$ is not reached}{
  Sample P uniformly from $[P_1, P_2]$\;
  Apply outlier detection for each cluster from $L$ to remove outliers from $D$\;
  \If{Number of texts in any cluster $n\ge P*N/K$}{
   Remove texts randomly from that cluster until $n\ge P*N/K$\;
   }
 Add the rest of $D$ to the train set and add all removed samples to the test set\;
 Train a classifier on the train set and update $L$ based on predictions of the classifier on the test set\;
 Calculate the criterion $\delta$ and update $j$;
 }
\caption{Enhancement of Clustering by the Iterative Classification}
\end{algorithm}

Averaged word embeddings were used as features in \cite{rakib2020enhancement,xu2017self}. One of the differences of our study is that we used USE representations\footnote{https://tfhub.dev/google/collections/universal-sentence-encoder/1} \cite{DBLP:journals/corr/abs-1803-11175,DBLP:journals/corr/abs-1907-04307} for short texts to plug them into one of the clustering algorithms: $k$-means, Hierarchical Agglomerative Clustering (HAC) or Spectral Clustering. We used a full similarity matrix as well as $k$-NN and similarity distribution based sparsification of the similarity matrix  \cite{rakib2018improving} with HAC. In both methods of sparsification, we set the number of non-zero elements in each row of the similarity matrix equal to the ratio of the number of samples in the dataset to the number of clusters. In addition, we tested all available linkage criteria for HAC. We tried the Isolation Forest (IF) \cite{liu2008isolation} and Local Outlier Factor (LOF) \cite{breunig2000lof} for outlier detection. We used clustering and outlier detection algorithms implemented in the scikit-learn\footnote{https://scikit-learn.org/stable/index.html} and scipy\footnote{https://www.scipy.org/} python libraries.

In contrast with \cite{rakib2020enhancement}, we used Transformer models such as BERT \cite{devlin2018bert} and RoBERTa \cite{liu2019roberta} for iterative fine-tuning and classification. In addition, we used Multinomial Logistic Regression (MLR) as in other works.

We consider two different stopping criteria. The first stopping criterion \cite{rakib2020enhancement} is defined as follows $\delta=\frac{1}{N}\sum_i|c_i-c^\prime_i|<\epsilon$ where $c_i$ and $c_i^\prime$ are sizes of clusters determined by the current labeling $L$ and previous labeling $L^\prime$, respectively, and $i$ is a cluster number. The second criterion is reached immediately when $\delta$ has a minimum value.

\section{Datasets}
Our study uses the same datasets as those in a number of previous studies \cite{xu2017self,hadifar2019self,rakib2020enhancement} on STC. The statistics on the datasets are presented in Table \ref{statistics}. The Search Snippets dataset is composed of Google search results. The texts in the Search Snippets dataset represent sets of key words, rather than being coherent texts. The Biomedical corpus is a subset of one of the BioAsQ\footnote{http://bioasq.org} challenge datasets. The texts in this dataset are paper titles with many special terms from biology and medicine. The Stack Overflow is a subset of the challenge on Kaggle and contains texts with question titles. AG News is a subset of the dataset that was used in \cite{zhang2015text}, where 2000 samples from each of the four categories were taken randomly. The Tweet, Google News TS, T and S sets are exactly those datasets which were used in \cite{yin2014dirichlet}. Note that the former and the latter four datasets can be grouped by the number of clusters. The first group contains relatively low numbers of clusters, while the second has greater numbers of clusters.  
\begin{table}
\centering
\begin{tabular}{llll}
\hline
\textbf{Dataset} & $K$ & $N$ & $M$ \\
\hline
Stack Overflow & 20 & 20000 & 8.2 \\
\hline
AG News & 4 & 8000 & 22.5 \\
\hline
Biomedical corpus & 20 & 20000 & 12.9 \\
\hline
Search Snippets & 8 & 12340 & 17.0 \\
\hline
Tweet & 89 & 2472 & 8.4 \\
\hline
Google News TS & 152 & 11109 & 28.0 \\
\hline
Google News T & 152 & 11109 & 6.2 \\
\hline
Google News S & 152 & 11109 & 21.8 \\
\hline

\end{tabular}
\caption{Statistics on the datasets used in the study. $K$ is the number of clusters, $N$ is the number of samples, $M$ is the average number of words in a document.}
\label{statistics}
\end{table}

\section{Results}
To measure the performance of our algorithm, we used such metrics as accuracy and Normalized Mutual Information (NMI). The value of NMI does not depend on the absolute values of labels. The accuracy is calculated using the Hungarian algorithm \cite{xu2017self}. It allows one to rearrange absolute label values to maximize accuracy.
\begin{table*}
\centering 
\begin{tabular}{llllll}
\hline
\textbf{Method} &  \textbf{Metric} & \textbf{Stack Overflow} & \textbf{AG News} & \textbf{Biomedical corpus} & \textbf{Search Snippets} \\
\hline
\multirow{2}{4em}{ECIC} & Acc. & 78.73$\pm$0.17 & 84.52$\pm$0.50 & 47.78$\pm$0.51 & \textbf{87.67$\pm$0.63} \\
& NMI &  73.44$\pm$0.35 & 59.07$\pm$0.84 & 41.27$\pm$0.36 & \textbf{71.93$\pm$1.04} \\
\hline
\multirow{2}{4em}{STA} & Acc. & 59.8$\pm$1.9 & - & \textbf{54.8$\pm$2.3} & 77.1$\pm$1.1 \\
& NMI & 54.8$\pm$1.0 & - & \textbf{47.1$\pm$0.8} & 56.7$\pm$1.0 \\
\hline
\multirow{2}{4em}{Init. clust.  $k$-means} & Acc. & \textbf{81.84$\pm$0.01} &  83.87$\pm$0.02 & 43.84$\pm$0.20 & 74.76$\pm$0.13 \\
& NMI & \textbf{80.80$\pm$0.01} & \textbf{61.88$\pm$0.04} & 37.85$\pm$0.13 & 54.25$\pm$0.16 \\
\hline
\multirow{2}{4em}{Iter. class. RoBERTa} & Acc. & \textbf{84.72$\pm$0.20} & \textbf{84.64$\pm$0.08} & 44.85$\pm$0.20 & 74.97$\pm$0.15 \\
& NMI & \textbf{80.63$\pm$0.97} & \textbf{62.69$\pm$0.20} & 38.40$\pm$0.13 & 55.17$\pm$0.26 \\
\hline
\multirow{2}{4em}{Iter. class. Log. Reg.} & Acc. & \textbf{83.31$\pm$0.05} & \textbf{86.53$\pm$0.1} & 44.96$\pm$0.17 & 75.87$\pm$0.15 \\
& NMI & \textbf{80.68$\pm$0.01} & \textbf{65.99$\pm$0.28} & 39.18$\pm$0.04 & 57.36$\pm$0.08 \\
\hline
\end{tabular}
\caption{Comparison with published results of accuracy and NMI scores for datasets with the smaller number of clusters.} 
\label{sota1}
\end{table*}

Our experiments on initial clustering tested which of the USE versions and which clustering algorithm should be used to obtain the best quality in terms of both aforementioned metrics. As a result, the old version of USE \cite{DBLP:journals/corr/abs-1803-11175} proved to be better (by a few percent) than the newer one \cite{DBLP:journals/corr/abs-1907-04307} in terms of both metrics on all 8 datasets. We tested $k$-means, HAC, and Spectral Clustering algorithms with these sentence embeddings. Interestingly, we found that the best clustering method was $k$-means for the whole group of datasets with the smaller number of clusters (see Table~\ref{sota1}). Since $k$-means is not a deterministic algorithm and its result depends on a particular initializatin, we averaged the results over 5 runs, each having 1000 initializations. On the contrary, HAC proved to be the best clustering method for datasets with the greater number of clusters (see Table~\ref{sota2}). Note we does not provide variance for HAC since this algorithm is determenistic.  Overall, $k$-NN sparsification  with the average linkage criterion gave the best results for the four datasets with the greater number of clusters. This differs from the results of \cite{rakib2020enhancement}, where a sparsification based on similarity distribution and the Ward linkage criterion are described as the most effective ones.

We obtained highly competitive results for two (Stack Overflow and AG News) of the four datasets from the first group of datasets. However, we did not get comparable results on the other two datasets (Search Snippets and Biomedical corpus), which can be easily explained. The Search Snippets dataset texts are sets of key words, rather than being coherent texts. Since USE was trained on coherent texts, it cannot produce a good result. The Biomedical dataset almost completely consists of special terms. USE probably did not see many of these terms during training, which explains its poor performance on this dataset. We got the best results for all four datasets from the second group in terms of NMI but not in terms of accuracy (see Table~\ref{sota2}).

\begin{table*}
\centering
\begin{tabular}{llllll}
\hline
\textbf{Method} &  \textbf{Metric} & \textbf{Tweet} & \textbf{Google News TS} & \textbf{Google News T} & \textbf{Google News S} \\
\hline
\multirow{2}{4em}{ECIC} & Acc. & \textbf{91.52$\pm$0.99} & \textbf{92.25$\pm$0.10} & \textbf{87.18$\pm$0.21} & \textbf{89.02$\pm$0.12} \\
& NMI &  86.87$\pm$0.13 & 94.40$\pm$0.11 & 87.87$\pm$1.00 & 89.96$\pm$0.11 \\
\hline
GSDPMM & NMI & 87.5$\pm$0.5 & 91.2$\pm$0.3 & 87.3$\pm$0.2 & 89.1$\pm$0.4 \\
\hline
\multirow{2}{4em}{Init. clust. HAC} & Acc. & 78.20 & 84.64 & 77.56 & 80.34 \\
& NMI & \textbf{91.28} & \textbf{94.77} & \textbf{91.14} & \textbf{91.96} \\
\hline
\end{tabular}
\caption{Comparison with published results of accuracy and NMI scores for datasets with the larger number of clusters.} 
\label{sota2}
\end{table*}

To improve the results of initial clustering, we tested the iterative classification algorithm with MLR and with neural pre-trained classifiers, such as BERT and RoBERTa. For the neural classifier, the number of iterations $T$ was set to be $10$, the learning rate $3\times 10^{-5}$ and the number of epochs to train during each iteration $2$. The use of the warm start i.e. training resumption after each iteration instead of re-initialization, and learning rate linear decaying schedule instead of the constant learning rate, did not show any considerable improvement. RoBERTa gave approximately one half percent improvement over the BERT performance. We set $T$ to be $50$ for MLR, since the algorithm worked more stable and had potential to improve for the more iterations than for neural classifiers. For the first stopping criterion we tried $\epsilon$ equal to $0.03$ and $0.05$. We found that the use of the second stopping criterion with neural classifiers gives better results than the first one. We did not use any criterion for MLR  and collected the metrics at the end of 50 iterations, since both considered metrics grew monotonically for this classifier. We set $P_1$ to be 0.75 and $P_2$ to be 0.95 for both types of classifiers. We averaged our results over 3 runs in both cases. We did not find any difference in the use of IF or LOF for outlier detection with all classifiers. 

The iterative classification achieved the state-of-the-art results on the Stack Overflow and AG News datasets with both types of classifiers and improved the good initial clustering result further (see Table \ref{sota1}). The neural classifier showed a one percent better performance for the Stack Overflow in terms of accuracy than MLR. We did not get comparable results for the Biomedical and Search Snippets datasets, since the iterative classification algorithm can improve the initial clustering result by a limited number of percent and it was low efficient for these two datasets. We did not observe any improvement for the second group of datasets, since it is more difficult for the algorithm to converge to the correct solution during iterations in the case of greater number of clusters.

\section{Conclusions}

The sentence embeddings based algorithm for enhanced clustering by iterative classification was applied to 8 datasets with short texts. The algorithm demonstrates state of the art results for the 6 out of 8 datasets. We argue that the lack of coherent and common texts causes an inferior performance of the algorithm for the two remaining datasets.

The quality of the whole algorithm strongly depends on the initial clustering quality. Initial clustering with USE representations has already allowed us to achieve a competitive performance for a number of datasets. Therefore, due to transfer learning these representations can be readily applied to other datasets even without iterative classification.

\bibliographystyle{acl_natbib}

\begin{thebibliography}{18}
\expandafter\ifx\csname natexlab\endcsname\relax\def\natexlab#1{#1}\fi

\bibitem[{Breunig et~al.(2000)Breunig, Kriegel, Ng, and
  Sander}]{breunig2000lof}
Markus~M Breunig, Hans-Peter Kriegel, Raymond~T Ng, and J{\"o}rg Sander. 2000.
\newblock Lof: identifying density-based local outliers.
\newblock In \emph{Proceedings of the 2000 ACM SIGMOD international conference
  on Management of data}, pages 93--104.

\bibitem[{Cer et~al.(2018)Cer, Yang, Kong, Hua, Limtiaco, John, Constant,
  Guajardo{-}Cespedes, Yuan, Tar, Sung, Strope, and
  Kurzweil}]{DBLP:journals/corr/abs-1803-11175}
Daniel Cer, Yinfei Yang, Sheng{-}yi Kong, Nan Hua, Nicole Limtiaco, Rhomni~St.
  John, Noah Constant, Mario Guajardo{-}Cespedes, Steve Yuan, Chris Tar,
  Yun{-}Hsuan Sung, Brian Strope, and Ray Kurzweil. 2018.
\newblock Universal sentence encoder.
\newblock \emph{CoRR}, abs/1803.11175.

\bibitem[{Devlin et~al.(2018)Devlin, Chang, Lee, and
  Toutanova}]{devlin2018bert}
Jacob Devlin, Ming-Wei Chang, Kenton Lee, and Kristina Toutanova. 2018.
\newblock Bert: Pre-training of deep bidirectional transformers for language
  understanding.
\newblock \emph{arXiv preprint arXiv:1810.04805}.

\bibitem[{Hadifar et~al.(2019)Hadifar, Sterckx, Demeester, and
  Develder}]{hadifar2019self}
Amir Hadifar, Lucas Sterckx, Thomas Demeester, and Chris Develder. 2019.
\newblock A self-training approach for short text clustering.
\newblock In \emph{Proceedings of the 4th Workshop on Representation Learning
  for NLP (RepL4NLP-2019)}, pages 194--199.

\bibitem[{Jipeng et~al.(2019)Jipeng, Zhenyu, Yun, Yunhao, and
  Xindong}]{jipeng2019short}
Qiang Jipeng, Qian Zhenyu, Li~Yun, Yuan Yunhao, and Wu~Xindong. 2019.
\newblock Short text topic modeling techniques, applications, and performance:
  a survey.
\newblock \emph{arXiv preprint arXiv:1904.07695}.

\bibitem[{Li et~al.(2017)Li, Duan, Wang, Zhang, Sun, and Ma}]{li2017enhancing}
Chenliang Li, Yu~Duan, Haoran Wang, Zhiqian Zhang, Aixin Sun, and Zongyang Ma.
  2017.
\newblock Enhancing topic modeling for short texts with auxiliary word
  embeddings.
\newblock \emph{ACM Transactions on Information Systems (TOIS)}, 36(2):1--30.

\bibitem[{Liu et~al.(2008)Liu, Ting, and Zhou}]{liu2008isolation}
Fei~Tony Liu, Kai~Ming Ting, and Zhi-Hua Zhou. 2008.
\newblock Isolation forest.
\newblock In \emph{2008 Eighth IEEE International Conference on Data Mining},
  pages 413--422. IEEE.

\bibitem[{Liu et~al.(2019)Liu, Ott, Goyal, Du, Joshi, Chen, Levy, Lewis,
  Zettlemoyer, and Stoyanov}]{liu2019roberta}
Yinhan Liu, Myle Ott, Naman Goyal, Jingfei Du, Mandar Joshi, Danqi Chen, Omer
  Levy, Mike Lewis, Luke Zettlemoyer, and Veselin Stoyanov. 2019.
\newblock Roberta: A robustly optimized bert pretraining approach.
\newblock \emph{arXiv preprint arXiv:1907.11692}.

\bibitem[{Nguyen et~al.(2015)Nguyen, Billingsley, Du, and
  Johnson}]{nguyen2015improving}
Dat~Quoc Nguyen, Richard Billingsley, Lan Du, and Mark Johnson. 2015.
\newblock Improving topic models with latent feature word representations.
\newblock \emph{Transactions of the Association for Computational Linguistics},
  3:299--313.

\bibitem[{Rakib et~al.(2018)Rakib, Jankowska, Zeh, and
  Milios}]{rakib2018improving}
Md~Rashadul~Hasan Rakib, Magdalena Jankowska, Norbert Zeh, and Evangelos
  Milios. 2018.
\newblock Improving short text clustering by similarity matrix sparsification.
\newblock In \emph{Proceedings of the ACM Symposium on Document Engineering
  2018}, pages 1--4.

\bibitem[{Rakib et~al.(2020)Rakib, Zeh, Jankowska, and
  Milios}]{rakib2020enhancement}
Md~Rashadul~Hasan Rakib, Norbert Zeh, Magdalena Jankowska, and Evangelos
  Milios. 2020.
\newblock Enhancement of short text clustering by iterative classification.
\newblock In \emph{Natural Language Processing and Information Systems}, pages
  105--117, Cham. Springer International Publishing.

\bibitem[{Xu et~al.(2015)Xu, Wang, Tian, Xu, Zhao, Wang, and Hao}]{xu2015short}
Jiaming Xu, Peng Wang, Guanhua Tian, Bo~Xu, Jun Zhao, Fangyuan Wang, and
  Hongwei Hao. 2015.
\newblock Short text clustering via convolutional neural networks.
\newblock In \emph{Proceedings of the 1st Workshop on Vector Space Modeling for
  Natural Language Processing}, pages 62--69.

\bibitem[{Xu et~al.(2017)Xu, Xu, Wang, Zheng, Tian, and Zhao}]{xu2017self}
Jiaming Xu, Bo~Xu, Peng Wang, Suncong Zheng, Guanhua Tian, and Jun Zhao. 2017.
\newblock Self-taught convolutional neural networks for short text clustering.
\newblock \emph{Neural Networks}, 88:22--31.

\bibitem[{Yang et~al.(2019)Yang, Cer, Ahmad, Guo, Law, Constant, {\'{A}}brego,
  Yuan, Tar, Sung, Strope, and Kurzweil}]{DBLP:journals/corr/abs-1907-04307}
Yinfei Yang, Daniel Cer, Amin Ahmad, Mandy Guo, Jax Law, Noah Constant,
  Gustavo~Hern{\'{a}}ndez {\'{A}}brego, Steve Yuan, Chris Tar, Yun{-}Hsuan
  Sung, Brian Strope, and Ray Kurzweil. 2019.
\newblock Multilingual universal
  sentence encoder for semantic retrieval.
\newblock \emph{CoRR}, abs/1907.04307.

\bibitem[{Yin and Wang(2014)}]{yin2014dirichlet}
Jianhua Yin and Jianyong Wang. 2014.
\newblock A dirichlet multinomial mixture model-based approach for short text
  clustering.
\newblock In \emph{Proceedings of the 20th ACM SIGKDD international conference
  on Knowledge discovery and data mining}, pages 233--242.

\bibitem[{Yin and Wang(2016)}]{yin2016model}
Jianhua Yin and Jianyong Wang. 2016.
\newblock A model-based approach for text clustering with outlier detection.
\newblock In \emph{2016 IEEE 32nd International Conference on Data Engineering
  (ICDE)}, pages 625--636. IEEE.

\bibitem[{Zhang et~al.(2019)Zhang, Dong, Yin, and Wang}]{zhang2019attentive}
Wei Zhang, Chao Dong, Jianhua Yin, and Jianyong Wang. 2019.
\newblock Attentive representation learning with adversarial training for short
  text clustering.
\newblock \emph{arXiv preprint arXiv:1912.03720}.

\bibitem[{Zhang and LeCun(2015)}]{zhang2015text}
Xiang Zhang and Yann LeCun. 2015.
\newblock Text understanding from scratch.
\newblock \emph{arXiv preprint arXiv:1502.01710}.

\end{thebibliography}

\end{document}